\documentclass[letterpaper, 10 pt, conference]{ieeeconf}  

\IEEEoverridecommandlockouts                              

\usepackage{color}
\usepackage{booktabs}
\usepackage{float}
\usepackage{diagbox}
\usepackage{graphicx}
\usepackage{cite}
\usepackage{changepage}
\usepackage{subfigure}
\usepackage{authblk}

\title{\LARGE \bf
A Driving Behavior Recognition Model with \\ Bi-LSTM and Multi-Scale CNN
}

\author{He Zhang, Zhixiong Nan*, Tao Yang, Yifan Liu and Nanning Zheng
\thanks{*Corresponding author: Zhixiong Nan{ \tt\small nzx2018@xjtu.edu.cn}}
\thanks{The authors are with the Institute of Artificial Intelligence and Robotics, Xi'an Jiaotong University, Xi’an, China}
}

\begin{document}

\maketitle
\thispagestyle{empty}
\pagestyle{empty}

\begin{abstract}

In autonomous driving, perceiving the driving behaviors of surrounding agents is important for the ego-vehicle to make a reasonable decision. In this paper, we propose a neural network model based on trajectories information for driving behavior recognition. Unlike existing trajectory-based methods that recognize the driving behavior using the hand-crafted features or directly encoding the trajectory, our model involves a Multi-Scale Convolutional Neural Network (MSCNN) module to automatically extract the high-level features which are supposed to encode the rich spatial and temporal information.  Given a trajectory sequence of an agent as the input, firstly, the Bi-directional Long Short Term Memory (Bi-LSTM) module and the MSCNN module respectively process the input, generating two features, and then the two features are fused to classify the behavior of the agent. We evaluate the proposed model on the public BLVD dataset, achieving a satisfying performance.

\end{abstract}
\section{INTRODUCTION}

Researches on understanding complex traffic scenarios have recently been widely studied in the autonomous driving community \cite{waltz1980understanding}. When constructing a safe and reliable autonomous driving system or Advanced Driver Assistance System (ADAS), in order to analyze the dynamic evolution of the traffic scene and then make a reasonable decision, it is necessary to perceive the driving behavior of other agents around the autonomous vehicle in real-time. For example, sensing the braking behavior and the lane changing behavior of vehicles in front of the autonomous vehicle is significant for predicting possible dangerous events. Meanwhile, accurate recognition of driving behavior can not only assist path planning and motion decisions but also serve as high-level semantics to assist trajectory prediction of vehicles or pedestrians \cite{deo2018multi,ding2019online}. In this paper, we focus on accurately identifying interactive behavior in the traffic environment, and the interactive behavior refers to the movement status of surrounding traffic agents (vehicles, pedestrians, riders, etc) relative to the ego-vehicle \cite{xue2019blvd}. The driving behavior categories of vehicles around the ego-vehicle are shown in Fig. \ref{Fig:behavior}.

In the autonomous driving environment, the trajectory sequence is considered as relatively reliable and valuable information to model traffic agent behaviors. Due to the complexity and dynamics of real traffic environments, it is challenging to classify the driving behavior. The main challenges are three-fold: 
\begin{figure}[htbp]
\begin{center}
\subfigure{
\begin{minipage}[t]{0.48\linewidth}
\centering
\includegraphics[width=1.5in]{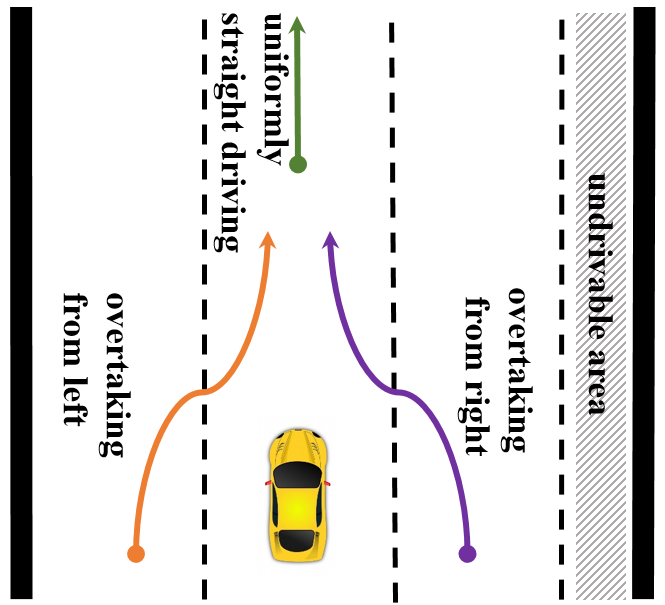}
\end{minipage}%
}%
\hfill
\subfigure{
\begin{minipage}[t]{0.48\linewidth}
\centering
\includegraphics[width=1.5in]{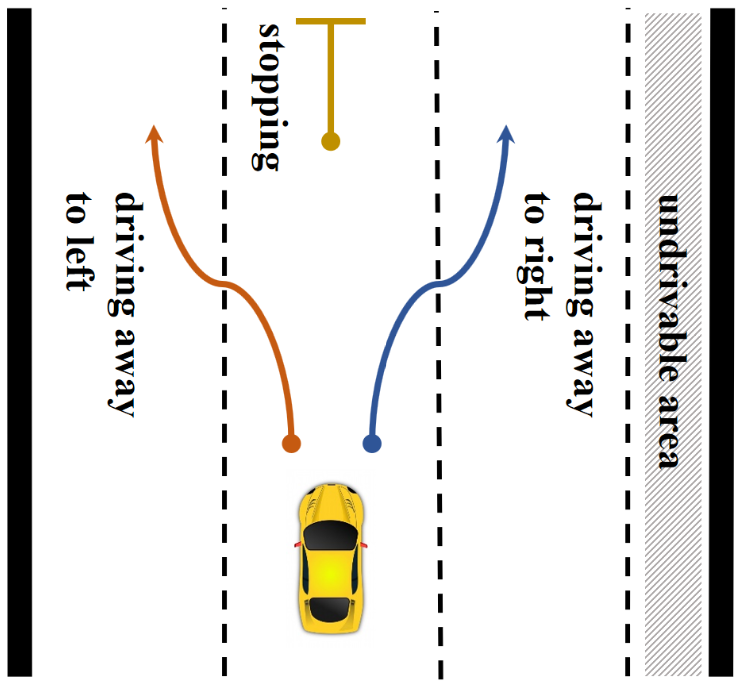}
\end{minipage}%
}%
\vspace{-0.2cm}
\subfigure{
\begin{minipage}[t]{0.48\linewidth}
\centering
\includegraphics[width=1.5in]{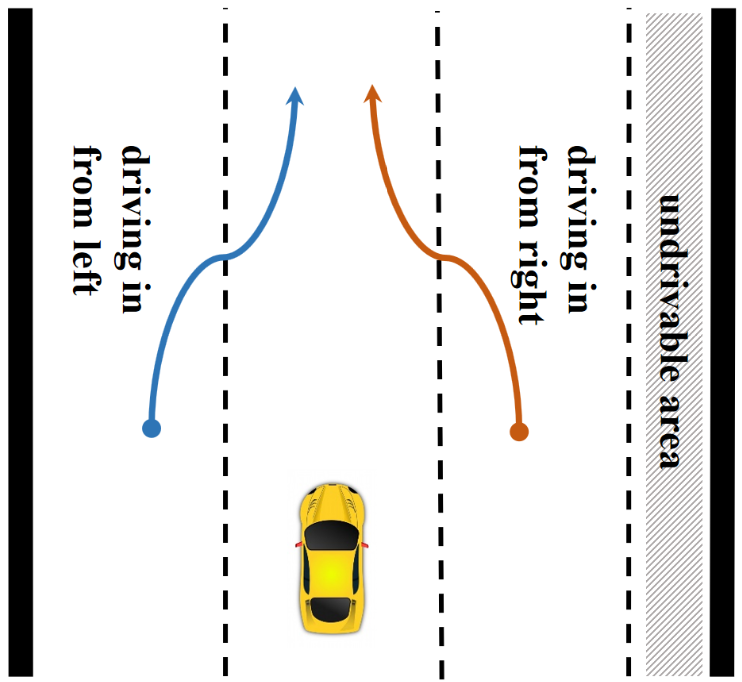}
\end{minipage}
}%
\subfigure{
\begin{minipage}[t]{0.48\linewidth}
\centering
\includegraphics[width=1.5in]{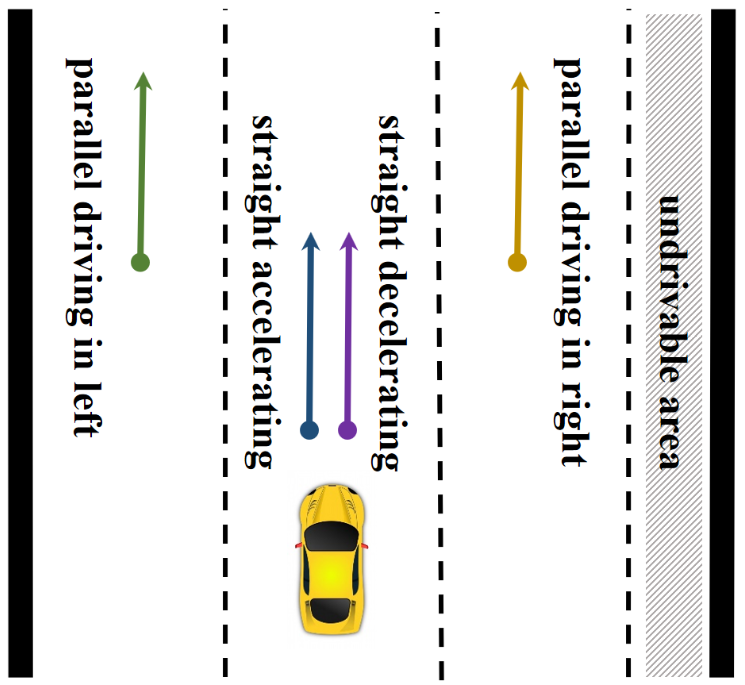}
\end{minipage}
}%
\end{center}
\vspace{-0.25cm}
\caption{Driving behavior categories}
\label{Fig:behavior}
\vspace{-0.4cm}
\end{figure}
1) Generally, each kind of driving event has different temporal durations. If we use a big window to split the trajectory into training samples, there may exist multiple kinds of behaviors in a sample; 2) For a fixed temporal window, the number and the behavior type of agents around ego-vehicle are highly dynamic; 3) There exists a severe imbalance in behavioral data, and the limited training samples are available for most anomalous behavior categories. 

Recent progress in Lidar, GPS and visual vehicle detection technologies allows collecting accurate and robust trajectory data, which makes it possible to leverage data-driven methods for driving behavior recognition task. Existing trajectory-based methods can be generally divided into two types, one is to construct a classifier based on some hand-crafted features\cite{svm1,svm2,hmm1, hmm2,hmm3,hmm4}, the other is to directly model the trajectory sequence to obtain dynamic evolution rules, and then implement the behavior classification\cite{hmm1-1,deo2018multi,ding2019online}. However, there exist many drawbacks for both of them. The former requires some domain knowledge to design manual features and generally needs to select different features for different datasets, leading to a lack of generalization across different scenarios. The drawback of the latter is that the original information included in trajectory points may be insufficient, which may lead to the under-fitting of the model. 

To overcome those drawbacks existing in the conventional methods, we propose a neural network model to recognize the driving behavior of surrounding agents. Unlike existing trajectory-based methods that recognize the driving behavior using the hand-crafted features or directly encoding the trajectory, our model first introduces a Multi-Scale Convolutional Neural Network (MSCNN) to automatically extract the high-level features. We note that the trajectory point used in our experiments includes four parameter channels, x, y, z coordinates and forward direction. On the one hand, with the powerful ability to process spatial information, the Convolutional Neural Network (CNN) architecture can detect potential local conjunctions among different parameter channels. This part of the information is generally ignored by the conventional sequence-based methods. On the other hand, the convolutional filters of different sizes can capture information of various time-scales. Consequently, the MSCNN module is supposed to encode richer spatial and temporal information than existing methods. Given a trajectory sequence of an agent as the input, we firstly utilize a Bi-directional Long Short Term Memory (Bi-LSTM) module to model the evolution rules of the trajectory, and then introduce the MSCNN module to generate the high-level features. Finally, the two features from the two modules are fused to classify the behavior of the agent. In addition, the Random Over-Sampling (ROS) module is leveraged during the training phase to handle the problem of the data imbalance. According to the experiment results on the public BLVD dataset \cite{xue2019blvd}, we can observe that our proposed method significantly outperforms other existing methods and the MSCNN module can improve the performance of all three metrics, with gains varying from 1.7\% to 2\%.

We summarize the contributions of this paper as follows:
\begin{itemize}
    \item This paper proposes a neural network model combining a Bi-LSTM module and a MSCNN module for the driving behavior classification task, and it is found that our model outperforms other existing methods on a large-scale dataset.
    \item We implement a classification framework with a data pre-processing module, enabling our method to efficiently alleviate the influence of data imbalance, which is significant for detecting anomalous events in traffic scenarios.
\end{itemize}

\section{RELATED WORK}

Trajectory-based driving behavior classification of traffic agents is an intensively-studied research topic, especially in the field of autonomous driving. Existing traffic behavior classification methods can be roughly divided into two categories according to the originally used information. One group of methods are feature-based algorithms, which deploy  features extracted from trajectory to build classifiers. The other group are sequence-based models, which directly model the trajectory sequences through HMM-based algorithms or neural networks.

Researchers have proposed many feature-based classifiers for vehicle behavior classification, including SVMs \cite{svm1,svm2}, HMMs \cite{hmm1, hmm2,hmm3,hmm4}, Naïve Bayes Classification \cite{dualapproach}, Logistic Regression \cite{dualapproach} and LSTM \cite{2016itsc}. These methods use motion-based features such as speed, acceleration, yaw rate and other context information such as lane position, turn signals, distance from the leading vehicle to implement the classifier. However, hand-crafted features design has a high demand for domain knowledge, and it also takes a lot of time to select different features combinations for different scenarios.

With the approach of the data explosion era, many data-driven methods focusing on the trajectory sequence have emerged to replace those feature-based methods, including HMM-based methods \cite{hmm1-1, hmm3}, neural network-based methods such as LSTM \cite{deo2018multi,ding2019online} and CNN \cite{cnn}. Some works take the behavior recognition as a branch to assist trajectory prediction at a high-level semantic level \cite{deo2018multi, ding2019online}.

Our proposed model is a kind of sequence-based method in a broad sense, without any manually-defined features as input. Unlike the previous sequence-based methods, to make full use of the information of the time dimension and the parameter channels, we first introduce a Multi-Scale CNN module to automatically extract the high-level features for the classification task.
   
\section{APPROACH}
      
\begin{figure*}[!htbp]
      \centering
      \vspace{0.075in}
      \includegraphics[scale=0.39]{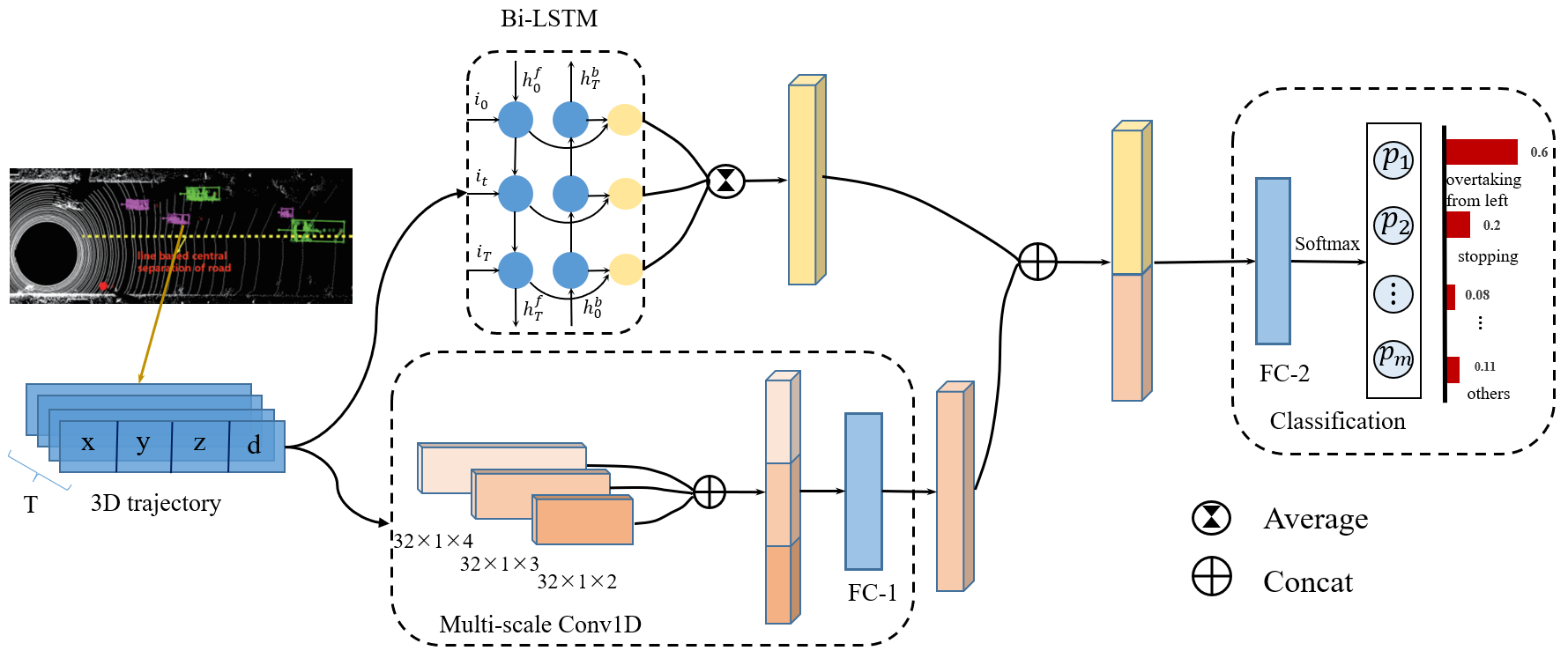}
      \vspace{-0.1cm}
      \caption{Firstly, given a trajectory sequence of an agent as the input, the Bi-directional Long Short Term Memory (Bi-LSTM) module and the MSCNN module respectively process the input, generating two features, then the two features are fused to classify the behavior of the agent.}
      \label{Fig:model}
      \vspace{-0.4cm}
   \end{figure*}

The classification framework can be divided into three parts, the data pre-processing part, the dynamic evolution rules modeling part and the high-level features extracting part. 
Firstly, to avoid our classifiers getting over-fitted to the classes with the majority of samples, we utilize a ROS module as the data pre-processing part to reduce the influence of the data imbalance, and this module is only implemented in the training phase. Secondly, for the dynamic evolution rules modeling part, a Bi-LSTM module is used to model the trajectory evolution rules. Thirdly, to increase the sufficiency of high-level representation, we introduce a MSCNN module for extracting multi-scale temporal features and spatial features. Finally, the two features of the two modules are fused to a fully connected layer (FC-2 in Fig. \ref{Fig:model}) to get the final classification results. The overview of the neural network model is summarized as Fig. \ref{Fig:model}.

\subsection{Problem Formulation} 

We formulate the driving behavior classification as a multi-class classification problem. For each traffic agent around an autonomous vehicle, we define its behavior category at each point according to its coordinates and driving status (accelerating, decelerating, etc.) relative to ego-vehicle. Following the formulation of driving event recognition in \cite{xue2019blvd}, we use 5 trajectory points near time $t$ to predict the behavior category at that time.

The input of the model is a state tensor of multiple trajectory points as

$$
{S}=\left[{s}^{\left(t-4\right)}, \ldots, {s}^{(t-1)}, {s}^{(t)}\right] \eqno{(1)}
$$
where,
$$
s^{(t)} = [x^{(t)}, y^{(t)},z^{(t)}, d^{(t)}] \eqno{(2)}
$$
are the $x$, $y$ and $z$ coordinates and direction information at time $t$ of the agent being predicted. All coordinates are the relative positions of some agent and ego-vehicle at time $ t $, and the direction $ d ^ {(t)} $ is the angle between each agent's forward direction and the centerline of the road.

We take the annotated behavior category $c^{(t)}$ corresponding to the last point $s^{(t)}$ as the true label of the sample $S$. The output of the model is the predicted behavior category $ \hat c ^ {(t)} $, as shown in 
$$
\hat c ^{(t)} = \mathop{\arg\max}_{i} P(i|s^{(t-4)},\cdots, s^{(t-1)},s^{t}) \eqno{(3)}
$$
where $P$ denotes the proposed classification model.

\subsection{Data Pre-processing} 

There exist some rarely occurred behaviors with extremely few available samples, leading to the severe data imbalance problem. The data imbalance easily increases the risk of a classification model overfitting to the classes with the majority of samples \cite{johnson2019survey}. To handle the problem of the data imbalance, we propose a data pre-processing module to balance the dataset. In this paper we focus on improving the performance of sequence-based methods, so we only explore several common data-level techniques: random under-sampling (RUS), random over-sampling (ROS), and weighted loss (WL) \cite{johnson2019survey, qin2018weighted}. After comparing the performance on the vehicle behavior dataset, we choose the simplest form the ROS method as the data pre-processing part, which is duplicating random samples from the minority group until the sample sizes of all classes are the same.

\subsection{Dynamic Evolution Rules Modeling} 

Recurrent neural networks have excellent performance in previous sequence-based methods, especially the LSTM architecture. Compared to vanilla LSTM, bidirectional LSTM (Bi-LSTM) architecture can use the bidirectional information of the trajectory and has better performance in text analysis tasks \cite{huang2015bidirectional}, which prompts us to introduce a Bi-LSTM module to model the dynamic evolution of trajectories. We average the output states of all moments to derive the final output of the Bi-LSTM module.

\subsection{High-Level Features Extracting} 

A crucial procedure of feature-based methods is to design efficient features such as speed and accelerating rate for building effective classifiers. But the design of manual features for some behavior requires specialized domain knowledge, and it is time-consuming to attempt different combinations of features for different behavior label spaces. To this end, it should be reasonable and efficient to use a neural network for automatically extracting spatial and temporal features. CNN is famous for its powerful ability to process spatial information, so we introduce this architecture with the aim of detecting potential local conjunctions among different parameter channels of the trajectory data. We notice that this part of the information is generally ignored by conventional sequence-based methods. Besides, to utilize features of various time scales, we design a MSCNN module with multiple convolution kernels of different sizes. Consequently, our MSCNN module is supposed to encode richer spatial and temporal information than existing methods.

Because the input sequence length in raw samples is 5, we choose three kinds of convolution kernels with size $1 \times 2$, $1 \times 3$ and $1 \times 4$ respectively to build the high-level feature extracting module. Then a fully connected layer (i.e. FC-1 in Fig. \ref{Fig:model}) is added to map the output of the MSCNN module to a lower dimension, aiming to control the information flow of this module.

\section{EXPERIMENT}

\subsection{Dataset} 

We evaluate the proposed model on the BLVD dataset \cite{xue2019blvd}, which is a large-scale 5D semantic benchmark set for traffic scenarios.
There are three types of traffic agents (vehicle, rider and pedestrian) in the dataset. Compared to other existing traffic datasets such as KITTI \cite{kitti} and Cityscape \cite{cordts2016cityscapes}, the BLVD benchmark is collected under more driving scenarios and various light conditions, with large-scale 3D trajectory annotation for 120K frames, making it a more practical and challenging dataset for our task. Moreover, we note that there are extremely fewer behavior categories and the limited scenes in most existing open-source driving behavior datasets, and the feature information varies from different datasets, which makes it difficult to integrate multiple data sources into a unified problem formulation. Thus we choose the BLVD dataset,  which has large-scale behavior annotations and abundant traffic scenarios, as the experimental platform to verify the effectiveness of our method.
Firstly, we filter out the trajectory sequences with length less than 7 and then split valid trajectories using a fixed window with size 5 to obtain samples we need. Then we remove the categories with less than 100 samples since these samples lack statistical significance. 
The sizes of the three kinds of agents are 142,442, 12,014, and 39,967, respectively. 
And there exists an imbalance problem in the behavior distribution of three kinds of agents, especially on the vehicle dataset. We split the filtered data as training and testing sets with a ratio of 8:2.

\subsection{Evaluation Metrics} 
Due to the high imbalance of the dataset, the accuracy rate cannot accurately reflect the true performance of the classifiers. Thus we introduce two metrics, balanced accuracy and F1-score, which are commonly used in machine learning for imbalanced multi-classification problems \cite{johnson2019survey}. In addition, we pay attention to whether some risky behaviors are detected, so recall is another evaluation criteria, which is also adopted by \cite{hmm4, kristoffersen2016towards} for trajectory-based behavior classification. We use $TP$, $TN$, $FP$ and $FN$ to denote true positive, true negative, false positive and false negative samples of predicted results, respectively.

\begin{itemize}
    \item Balanced Accuracy
    $$
    Balanced\,Accuracy=\frac{1}{2}\times(TPR+TNR)\eqno{(4)}
    $$
    
 where TPR denotes $TP/(TP+FP)$, TNR denotes $TN/(TN+FP)$.
    
    \item Recall
    $$
    Recall = \frac{TP}{TP+FN} \eqno{(5)}
    $$
    \item F1-score
    $$
    F1\rule[3pt]{5pt}{0.5pt} score= \frac{(1+\beta^{2})\times Recall\times Precision}{\beta^{2}\times Recall + Precision }\eqno{(6)}
    $$
    
where Precision denotes $TP/(TP+FP)$, and the coefficient $\beta$ is used to adjust the relative importance of precision versus recall. We set $\beta$ to 1 in this paper.

\end{itemize}

\begin{table*}[]
\vspace{0.075in}
\caption{\label{tab: test} Average performance of all behavior categories on three datasets. For a fair comparison, the ROS module is implemented in all methods}
\resizebox{\textwidth}{15mm}{
\begin{tabular}{cccccccccc}
\toprule
Dataset      & \multicolumn{3}{c}{Vehicle}            & \multicolumn{3}{c}{Pedestrian}         & \multicolumn{3}{c}{Rider}              \\
\cmidrule(r){1-1} \cmidrule(r){2-4} \cmidrule(r){5-7} \cmidrule(r){8-10}
\diagbox{Model}{Metric} & Balanced Accuracy & F1-score & Recall  & Balanced Accuracy & F1-score & Recall  & Balanced Accuracy & F1-score & Recall  \\
\midrule
HMM          & 51.51\%           & 38.49\%  & 58.14\% & 86.77\%           & 81.93\%  & 90.77\% & 66.47\%           & 65.18\%  & 69.45\% \\

Conv1D \cite{cnn}          & 86.55\%           & 79.01\%  & 88.07\% & 97.84\%           & 97.38\%  & \textbf{98.15\%} & 90.84\%           & 90.72\%  & 91.47\% \\

LSTM \cite{2016itsc}        & 86.36\%           & 81.54\%  & 85.76\% & 97.51\%           & 96.89\%  & 97.75\% & 93.49\%           & 93.32\%  & 93.91\% \\

Our Method   & \textbf{90.85\%}           & \textbf{88.07\%}  & \textbf{90.27\%} & \textbf{97.89\%}           &\textbf{97.75\%}  & 97.77\% & \textbf{95.00\%}           & \textbf{94.97\%}  & \textbf{95.09\%} \\
\bottomrule
\end{tabular}}
\vspace{-0.3cm}
\end{table*}

\subsection{Implementation Details}

In the Bi-LSTM module, we set 2 and 64 as the number of LSTM layers and the number of hidden states. In the MSCNN module, each kind of convolution kernels has 32 different channels, and the output dimension of FC-2 is set to 32. All the parameters mentioned above are obtained based on cross-validation. The numbers of output classes of three kinds of agents are 13, 6 and 7, respectively.
In the training phase, all neural networks are trained for 60 epochs with batch size 256. The initial learning rate is set to 0.005, then decays to 0.001 after 40 epochs. Moreover, cross-entropy and Adam \cite{kingma2014adam} are adopted as the loss function and optimizer, respectively.    
The HMM model is implemented by Scikit-learn \cite{pedregosa2011scikit}, and the other three models are implemented with a deep learning framework Pytorch \cite{paszke2019pytorch}.

\subsection{Experiment Design and Results Analysis}

\textbf{Compare with baselines.}
The proposed model is a sequence-based method in a broad sense. And we introduce a MSCNN module that can automatically extract the features of the time dimension and the parameter channels to provide more valuable information to the classifier. To verify the effectiveness of our idea, we compare our model with three baseline methods, which are briefly introduced as follows.

\begin{itemize}
    \item HMM: For each unique behavioral class, an HMM model is trained using the Baum Welch algorithm based on the corresponding trajectory sequences. Because the observation states (3D trajectory cues) are multidimensional continuous variables, we choose hidden Markov models with gaussian emissions, and the number of hidden states of each HMM model is set to 7.   
    \item Conv1D : Mammeri et al \cite{cnn} developed the Conv1D model for vehicle maneuver classification, which applies 4 convolutional layers for extracting features and a fully connected layer as the classification module. 
    \item LSTM : A vanilla LSTM model is implemented for behavior classification using 3D trajectory cues \cite{2016itsc,deo2018multi,ding2019online}, with 2 layers and 64-dimension hidden states.  
\end{itemize}

The average performance of baseline methods and the proposed method on three metrics are summarized in Tab. \ref{tab: test}. Overall, the performance of the proposed model is significantly better than several baseline methods. It is apparent that the performances on the vehicle dataset are worse than those on the other two datasets. A possible reason is that the vehicle dataset has a more severe imbalance and more categories compared to the other two datasets, bringing great difficulty for behavior classification. 
\begin{table}[]
\caption{Results of the ablative experiment}
\label{tab:ablation}
\begin{tabular*}{\hsize}{@{}@{\extracolsep{\fill}}cccc@{}}
\toprule
\diagbox{Model}{Metric}      & Balanced Accuracy & F1-score & Recall   \\
\midrule
Bi-LSTM           & 85.27\%           & 82.55\%  & 85.27\%  \\
Bi-LSTM+MSCNN     & 85.73\%           & 83.03\%  & 85.73\%  \\
ROS+Bi-LSTM       & 88.92\%           & 86.20\%  & 89.18\%  \\
ROS+Bi-LSTM+MSCNN & \textbf{90.85\%}           & \textbf{88.07\%}  & \textbf{90.27\%}  \\
\bottomrule
\end{tabular*}
\vspace{-0.5cm}
\end{table}

Proposed method: As is shown in Tab. \ref{tab: test}, the proposed model outperforms other methods on all datasets. Especially on the vehicle dataset, our method achieves 4.30\%, 6.53\% and 2.20\% improvement over the second-best methods respectively on the three metrics. Our method outperforms others for two reasons: 1) Compared to HMM and LSTM methods that focus on modeling the dynamic laws of trajectories, we innovatively introduce a MSCNN module to automatically extract multi-scale features, which incorporates abundant spatial and temporal information into our model; 2)  Compared to Conv1D method that encodes the kinetic patterns with a few convolutional layers, the Bi-LSTM branch of our model has stronger capacity in learning the dynamic characteristics of agents’ behaviors. 
In addition, the performance of the proposed model is similar to that of other models on the pedestrian dataset, which shows that the behavioral pattern of pedestrians is easier to learn.


Baselines: The HMM model performs worst among the three baselines. To ensure that the original input of all models is consistent, the observation sequence length of the HMM model is also set to 5. Therefore, the short sequence length may be the main reason causing the poor performance of HMMs. The other two baselines perform similarly on the vehicle and pedestrian datasets,
\begin{figure}[!htbp]
      \centering
      \vspace{0.3cm}
      \includegraphics[scale=0.24]{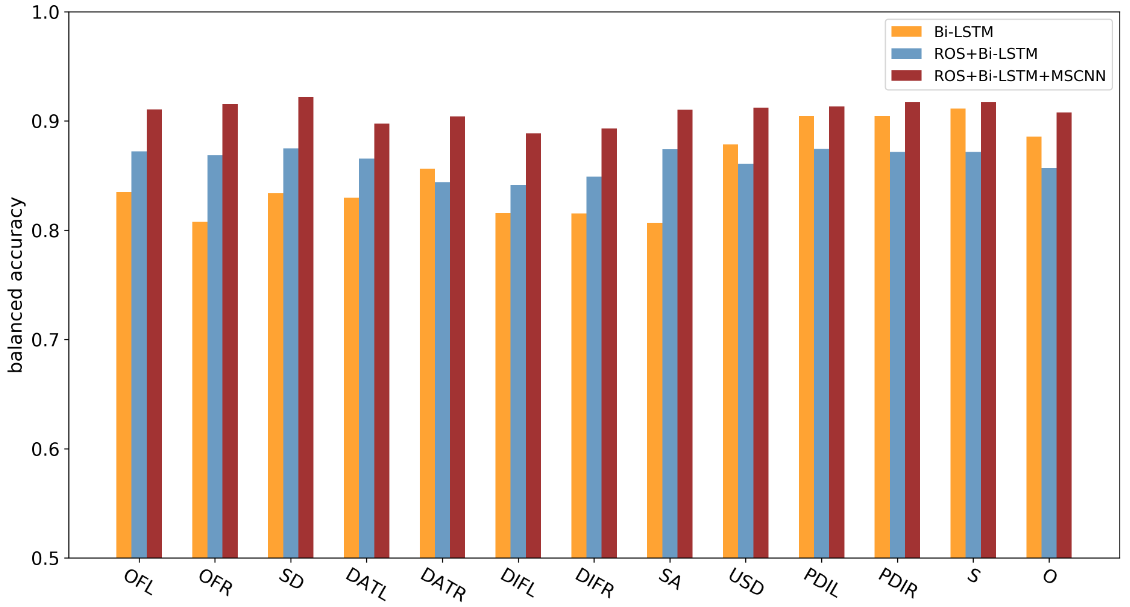}
      \vspace{-0.2cm}
      \caption{Performance of all behavior categories on vehicle dataset}
      \label{Fig:balance}
      \vspace{-0.4cm}
   \end{figure}
and the Conv1D model performs better on the rider dataset, which shows that the Conv1D model can capture more movement patterns of riders. 
   
\textbf{Effectiveness of MSCNN and ROS.}
We design an ablative experiment to explore the influence of each component in our method. The detailed results are reported in Tab. \ref{tab:ablation}. We also report the balanced accuracy of all behavior categories in the vehicle dataset in Fig. \ref{Fig:balance}, and to get better visualization we use the abbreviation of behavior category on the horizontal axis of this figure, where OFL, OFR, SD, DATL, DATR, DIFL, DIFR, SA, USD, PDIL, PDIR, S, O  correspond to \textit{overtaking from left}, \textit{overtaking from right}, \textit{straight decelerating}, \textit{driving away to left}, \textit{driving away to right}, \textit{driving in from left}, \textit{driving in from right},  \textit{straight accelerating}, \textit{uniformly straight driving}, \textit{parallel driving in left}, \textit{parallel driving in right}, \textit{stopping}, \textit{others}, respectively. 

If the ROS module is added, we obtain significant performance improvement on three metrics. As shown in Fig. \ref{Fig:balance}, it is apparent that the ROS module obviously improves the performance of most minority behavior categories, which proves that the ROS module can efficiently reduce the impact of the data imbalance. And we note that most of the behaviors in minority categories are anomalous events in actual traffic scenarios, such as overtaking, decelerating straight, etc. Therefore the performance improvement of these minority categories has good practical significance.

\begin{figure}[!htbp]
      \centering
      \vspace{0.075in}
      \includegraphics[scale=0.18]{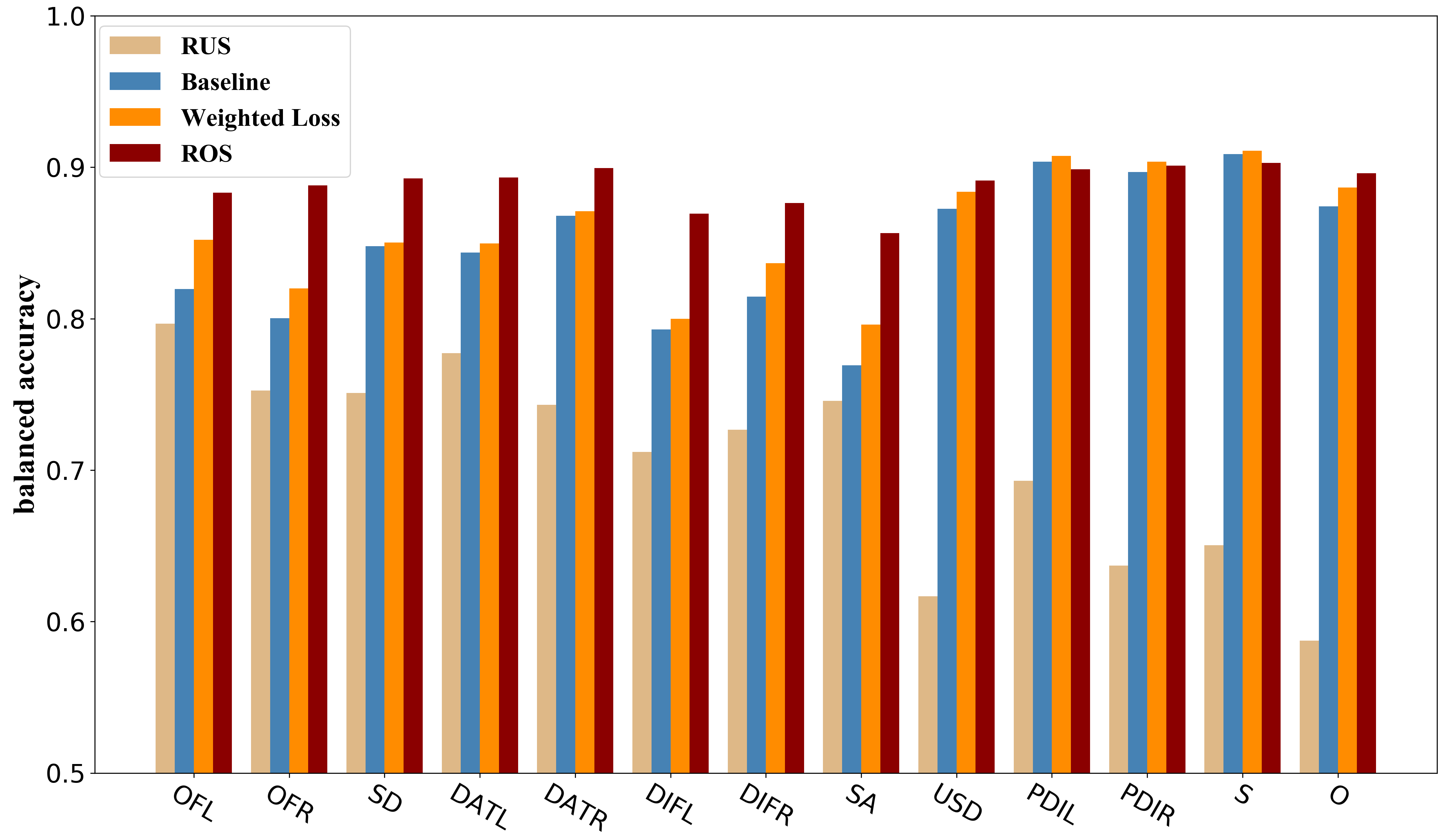}
      \vspace{-0.1cm}
      \caption{Comparison results of different data pre-pocessing methods on vehicle dataset. The behavior type indexes in x and y axis are the same as ones in Fig. \ref{Fig:balance}}
      \label{Fig:process}
      \vspace{-0.4cm}
   \end{figure}

If the MSCNN module is added, we find the performance of three metrics is also improved, with gains varying from 1.7\% to 2\%, which proves that the MSCNN module can introduce extra information for assisting the classification. Moreover, according to the performance of the Bi-LSTM+MSCNN model and the ROS+Bi-LSTM+MSCNN model, the ROS module can assist the MSCNN module to achieve more significant improvement, apparently. This is because the MSCNN module can pay more attention to the minority classes when the influence of data imbalance is alleviated.

The comparison results of different data pre-processing methods are shown in Fig. \ref{Fig:process}, where \textit{Baseline} denotes the proposed model without the data pre-processing module. It is apparent that ROS significantly outperforms other methods, so we incorporate this method into our learning framework.

\begin{figure}[!htbp]
      \centering
      \vspace{0.075in}
      \includegraphics[scale=0.32]{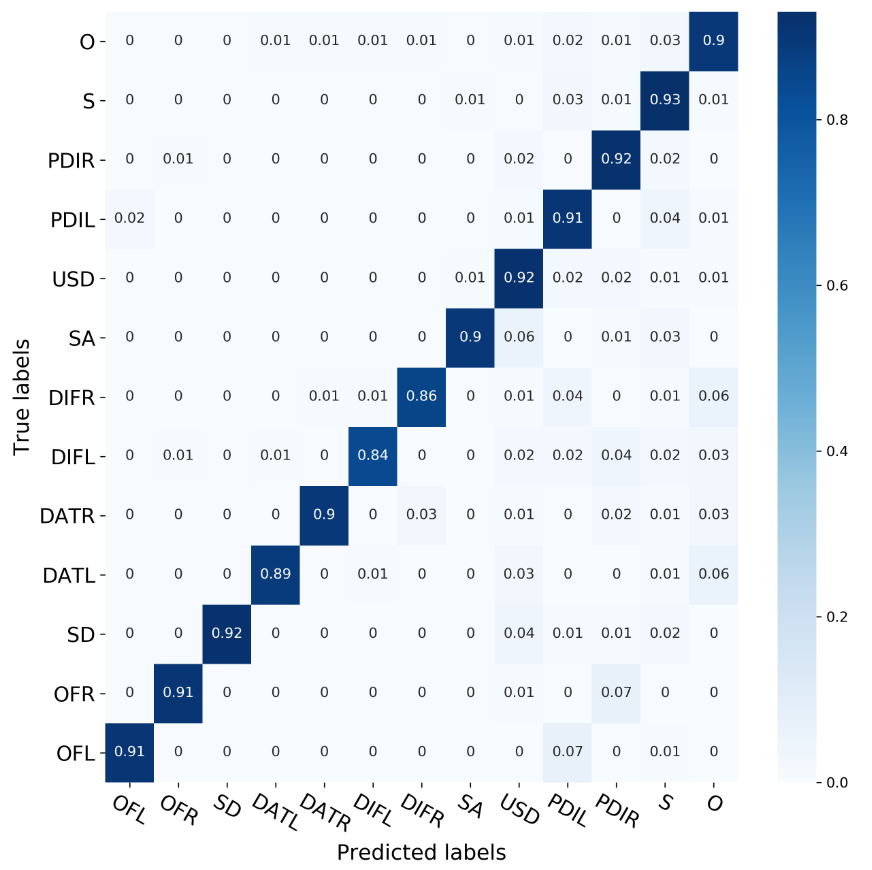}
      \vspace{-0.1cm}
      \caption{Confusion matrix of predicted results on the vehicle dataset. The behavior type indexes in x and y axis are the same as ones in Fig. \ref{Fig:balance}}
      \label{Fig:confusion}
      \vspace{-0.4cm}
   \end{figure}

\textbf{Error Analysis.}
Fig. \ref{Fig:confusion} shows the confusion matrix of the classification results on the vehicle dataset, and the elements on the diagonal represent the recall rate of each type of behavior. As shown in Fig. \ref{Fig:confusion}, there are two obvious false positive (FP) rates, one is that OFL
is wrongly predicted into PDIL, and the other is that OFR is wrongly predicted into PDIR. These two pairs of behaviors are very similar in the physical sense, and there may exist a partial overlap between two kinds of behaviors.

When using a fixed time window with a small length (e.g. 5) to split the trajectory sequences, samples from this overlap interval will bring great difficulty to classification, which affects the improvement of our method. A natural idea to avoid this issue is to increase the length of the time window, which may bring invalid samples with multiple behaviors, as mentioned in section III. In future work, improving the performance by introducing the bigger time window is a potential research topic. And we need to trade-off between simplifying problem formulation and pursuing the extreme performance of the model.

\section{CONCLUSIONS}

In this paper, we propose a neural network model for driving behavior recognition. Compared with the existing sequence-based methods, we introduce a MSCNN module that can automatically extract the features of the time dimension and the parameter channels, providing more valuable information to the classifier. By comparing our method with several existing methods on the BLVD dataset and analyzing the ablative experiment, we draw the conclusion that 1) our method outperforms all other methods, 2) the MSCNN module enables to encode the abundant spatial and temporal information to assist classification, 3) the ROS module has a significant impact in eliminating the imbalance in the trajectory data.

\section{ACKNOWLEDGEMENT}

This work is supported by the National Science Foundation of China (No. 61790562, 61790563, 61773312)


\addtolength{\textheight}{-12cm}   





\bibliographystyle{ieeetr}
\bibliography{reference}

\begin{thebibliography}{10}

\bibitem{waltz1980understanding}
D.~L. Waltz, ``Understanding scene descriptions as event simulations,'' in {\em
  Proceedings of the 18th annual meeting on Association for Computational
  Linguistics}, pp.~7--11, Association for Computational Linguistics, 1980.

\bibitem{deo2018multi}
N.~Deo and M.~M. Trivedi, ``Multi-modal trajectory prediction of surrounding
  vehicles with maneuver based lstms,'' in {\em 2018 IEEE Intelligent Vehicles
  Symposium (IV)}, pp.~1179--1184, IEEE, 2018.

\bibitem{ding2019online}
W.~Ding and S.~Shen, ``Online vehicle trajectory prediction using policy
  anticipation network and optimization-based context reasoning,'' {\em arXiv
  preprint arXiv:1903.00847}, 2019.

\bibitem{xue2019blvd}
J.~Xue, J.~Fang, T.~Li, B.~Zhang, P.~Zhang, Z.~Ye, and J.~Dou, ``Blvd: Building
  a large-scale 5d semantics benchmark for autonomous driving,'' {\em arXiv
  preprint arXiv:1903.06405}, 2019.

\bibitem{svm1}
H.~M. Mandalia and M.~D.~D. Salvucci, ``Using support vector machines for
  lane-change detection,'' in {\em Proceedings of the human factors and
  ergonomics society annual meeting}, vol.~49, pp.~1965--1969, SAGE
  Publications Sage CA: Los Angeles, CA, 2005.

\bibitem{svm2}
G.~S. Aoude, B.~D. Luders, K.~K. Lee, D.~S. Levine, and J.~P. How, ``Threat
  assessment design for driver assistance system at intersections,'' in {\em
  13th International IEEE Conference on Intelligent Transportation Systems},
  pp.~1855--1862, IEEE, 2010.

\bibitem{hmm1}
C.~Laugier, I.~E. Paromtchik, M.~Perrollaz, M.~Yong, J.-D. Yoder, C.~Tay,
  K.~Mekhnacha, and A.~N{\`e}gre, ``Probabilistic analysis of dynamic scenes
  and collision risks assessment to improve driving safety,'' {\em IEEE
  Intelligent Transportation Systems Magazine}, vol.~3, no.~4, pp.~4--19, 2011.

\bibitem{hmm2}
B.~T. Morris and M.~M. Trivedi, ``Trajectory learning for activity
  understanding: Unsupervised, multilevel, and long-term adaptive approach,''
  {\em IEEE transactions on pattern analysis and machine intelligence},
  vol.~33, no.~11, pp.~2287--2301, 2011.

\bibitem{hmm3}
H.~Berndt, J.~Emmert, and K.~Dietmayer, ``Continuous driver intention
  recognition with hidden markov models,'' in {\em 2008 11th International IEEE
  Conference on Intelligent Transportation Systems}, pp.~1189--1194, IEEE,
  2008.

\bibitem{hmm4}
J.~V. Dueholm, M.~S. Kristoffersen, R.~K. Satzoda, T.~B. Moeslund, and M.~M.
  Trivedi, ``Trajectories and maneuvers of surrounding vehicles with panoramic
  camera arrays,'' {\em IEEE Transactions on Intelligent Vehicles}, vol.~1,
  no.~2, pp.~203--214, 2016.

\bibitem{hmm1-1}
N.~Deo, A.~Rangesh, and M.~M. Trivedi, ``How would surround vehicles move? a
  unified framework for maneuver classification and motion prediction,'' {\em
  IEEE Transactions on Intelligent Vehicles}, vol.~3, no.~2, pp.~129--140,
  2018.

\bibitem{dualapproach}
A.~Sonka, F.~Krauns, R.~Henze, F.~K{\"u}{\c{c}}{\"u}kay, R.~Katz, and U.~Lages,
  ``Dual approach for maneuver classification in vehicle environment data,'' in
  {\em 2017 IEEE Intelligent Vehicles Symposium (IV)}, pp.~97--102, IEEE, 2017.

\bibitem{2016itsc}
A.~Khosroshahi, E.~Ohn-Bar, and M.~M. Trivedi, ``Surround vehicles trajectory
  analysis with recurrent neural networks,'' in {\em 2016 IEEE 19th
  International Conference on Intelligent Transportation Systems (ITSC)},
  pp.~2267--2272, IEEE, 2016.

\bibitem{cnn}
A.~Mammeri, Y.~Zhao, A.~Boukerche, A.~J. Siddiqui, and B.~Pekilis, ``Design of
  a semi-supervised learning strategy based on convolutional neural network for
  vehicle maneuver classification,'' in {\em 2019 IEEE International Conference
  on Wireless for Space and Extreme Environments (WiSEE)}, pp.~65--70, IEEE,
  2019.

\bibitem{johnson2019survey}
J.~M. Johnson and T.~M. Khoshgoftaar, ``Survey on deep learning with class
  imbalance,'' {\em Journal of Big Data}, vol.~6, no.~1, p.~27, 2019.

\bibitem{qin2018weighted}
R.~Qin, K.~Qiao, L.~Wang, L.~Zeng, J.~Chen, and B.~Yan, ``Weighted focal loss:
  An effective loss function to overcome unbalance problem of chest x-ray14,''
  in {\em IOP Conference Series: Materials Science and Engineering}, vol.~428,
  p.~012022, IOP Publishing, 2018.

\bibitem{huang2015bidirectional}
Z.~Huang, W.~Xu, and K.~Yu, ``Bidirectional lstm-crf models for sequence
  tagging,'' {\em arXiv preprint arXiv:1508.01991}, 2015.

\bibitem{kitti}
A.~Geiger, P.~Lenz, C.~Stiller, and R.~Urtasun, ``Vision meets robotics: The
  kitti dataset,'' {\em The International Journal of Robotics Research},
  vol.~32, no.~11, pp.~1231--1237, 2013.

\bibitem{cordts2016cityscapes}
M.~Cordts, M.~Omran, S.~Ramos, T.~Rehfeld, M.~Enzweiler, R.~Benenson,
  U.~Franke, S.~Roth, and B.~Schiele, ``The cityscapes dataset for semantic
  urban scene understanding,'' in {\em Proceedings of the IEEE conference on
  computer vision and pattern recognition}, pp.~3213--3223, 2016.

\bibitem{kristoffersen2016towards}
M.~S. Kristoffersen, J.~V. Dueholm, R.~K. Satzoda, M.~M. Trivedi, A.~Mogelmose,
  and T.~B. Moeslund, ``Towards semantic understanding of surrounding vehicular
  maneuvers: A panoramic vision-based framework for real-world highway
  studies,'' in {\em Proceedings of the IEEE Conference on Computer Vision and
  Pattern Recognition Workshops}, pp.~41--48, 2016.

\bibitem{kingma2014adam}
D.~P. Kingma and J.~Ba, ``Adam: A method for stochastic optimization,'' {\em
  arXiv preprint arXiv:1412.6980}, 2014.

\bibitem{pedregosa2011scikit}
F.~Pedregosa, G.~Varoquaux, A.~Gramfort, V.~Michel, B.~Thirion, O.~Grisel,
  M.~Blondel, P.~Prettenhofer, R.~Weiss, V.~Dubourg, {\em et~al.},
  ``Scikit-learn: Machine learning in python,'' {\em Journal of machine
  learning research}, vol.~12, no.~Oct, pp.~2825--2830, 2011.

\bibitem{paszke2019pytorch}
A.~Paszke, S.~Gross, F.~Massa, A.~Lerer, J.~Bradbury, G.~Chanan, T.~Killeen,
  Z.~Lin, N.~Gimelshein, L.~Antiga, {\em et~al.}, ``Pytorch: An imperative
  style, high-performance deep learning library,'' in {\em Advances in Neural
  Information Processing Systems}, pp.~8024--8035, 2019.

\end{thebibliography}

\end{document}